\def\cvprPaperID{xxxx}
\def\confName{CVPR}
\def\confYear{xxxx}

\def\paperTitle{Human-to-Human Interaction Detection}

\def\authorBlock{
    Zhenhua Wang$^\dagger$ \qquad
    Kaining Ying\textsuperscript{$\ddagger$}  \qquad
    Jiajun Meng\textsuperscript{$\ddagger$}  \qquad
    Jifeng Ning\textsuperscript{$\dagger$}  
    \\

    $^\dagger$Intelligent Media Processing Group, College of Information Engineering, Northwest A\&F University\\
    $^\ddagger $College of Computer Science and Technology, Zhejiang University of Technology\\

    {\tt\small kaining.ying.cv@gmail.com}
}

\newif\ifreview 
\newif\ifarxiv \newcommand{\arxiv}{\arxivtrue}
\newif\ifcamera 
\newif\ifrebuttal

\arxiv

\pdfoutput=1
\documentclass[10pt,twocolumn,letterpaper]{article}
\ifreview \usepackage[review]{cvpr} \fi
\ifarxiv \usepackage[pagenumbers]{cvpr} \fi
\ifrebuttal \usepackage[rebuttal]{cvpr} \fi
\ifcamera \usepackage{cvpr} \fi

\usepackage{graphicx}
\usepackage{amsmath, bm}
\usepackage{amssymb}
\usepackage{booktabs}
\usepackage{bigstrut}
\usepackage{threeparttable} 
\usepackage{bbding}
\usepackage[misc]{ifsym}

\usepackage[linesnumbered,ruled]{algorithm2e}

\usepackage{times}
\usepackage{microtype}
\usepackage{epsfig}
\usepackage[table,xcdraw]{xcolor}
\usepackage{caption}
\usepackage{float}
\usepackage{placeins}
\usepackage{color, colortbl}
\usepackage{stfloats}
\usepackage{enumitem}
\usepackage{tabularx}
\usepackage{xstring}
\usepackage{multirow}
\usepackage{xspace}
\usepackage{url}
\usepackage{subcaption}
\usepackage{xcolor}
\usepackage[hang,flushmargin]{footmisc}
\ifcamera \usepackage[accsupp]{axessibility} \fi

% ...

\ifarxiv  \fi

\newcommand{\R}[1]{{%
    \textbf{%
        \ifstrequal{#1}{1}{\textcolor{red}{R#1}}{%
        \ifstrequal{#1}{2}{\textcolor{blue}{R#1}}{%
        \ifstrequal{#1}{3}{\textcolor{magenta}{R#1}}{%
        \ifstrequal{#1}{4}{\textcolor{teal}{R#1}}{%
                           \textcolor{cyan}{R#1}%
        }}}}%
    }%
}}

\usepackage{xr-hyper}

\makeatletter
\newcommand*{\addFileDependency}[1]{
  \typeout{(#1)}
  \@addtofilelist{#1}
  \IfFileExists{#1}{}{\typeout{No file #1.}}
}

\makeatother

\usepackage[pagebackref,breaklinks,colorlinks]{hyperref}
\usepackage[capitalize]{cleveref}
\crefname{section}{Sec.}{Secs.}
\crefname{table}{Table}{Tables}
\crefname{figure}{Fig.}{Figs.}

\begin{document}
%% TITLE
\title{\paperTitle}
\author{\authorBlock}

\makeatletter
\let\@oldmaketitle\@maketitle% Store \@maketitle
\renewcommand{\@maketitle}{\@oldmaketitle% Update \@maketitle to insert...
 \centering
    \includegraphics[width=\linewidth]{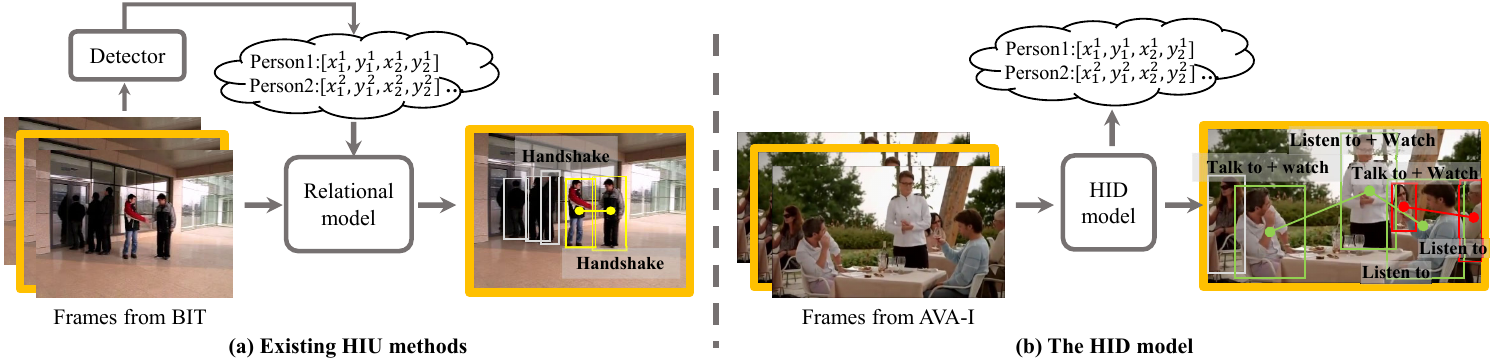}
    \captionof{figure}{Human interaction understanding (HIU) with existing methods (left) and the proposed 
    human-to-human interaction detection (HID) model (right). 
    \textbf{(a)} The former typically involves a human detection stage and a relational modeling stage, 
    and these two stages are trained separately \cite{cagnet, arg, gdn}. Moreover, only simple and choreographed benchmarks (\eg \emph{BIT} \cite{bit}, \emph{UT} \cite{ut} and \emph{TVHI} \cite{tvhi}) are available, which constrains the development of HIU techniques. \textbf{(b)} We address these issues by proposing the HID task and a challenging benchmark (namely AVA-I) of daily human interactions. We also present a strong baseline which enables the prediction of human-to-human interactions in a single stage.
    }
    \label{fig:teaser}
    \bigskip}                   % ... an image
\makeatother

\maketitle
%%
% DONE:
\begin{abstract}
A comprehensive understanding of interested human-to-human interactions in video streams, such as queuing, handshaking, fighting and chasing, is of immense importance to the surveillance of public security in regions like campuses, squares and parks. Different from conventional human interaction recognition, which uses choreographed videos as inputs, neglects concurrent interactive groups, and performs detection and recognition in separate stages, we introduce a new task named \textbf{human-to-human interaction detection (HID)}. HID devotes to detecting subjects, recognizing person-wise actions, and grouping people according to their interactive relations, in one model. First, based on the popular AVA dataset created for action detection, we establish a new HID benchmark, termed \textbf{AVA-Interaction (AVA-I)}, by adding annotations on interactive relations in a frame-by-frame manner. AVA-I consists of 85,254 frames and 86,338 interactive groups, and each image includes up to 4 concurrent interactive groups. Second, we present a novel baseline approach \textbf{SaMFormer} for HID, containing a visual feature extractor, a split stage which leverages a Transformer-based model to decode action instances and interactive groups, and a merging stage which reconstructs the relationship between instances and groups. All SaMFormer components are jointly trained in an end-to-end manner. Extensive experiments on AVA-I validate the superiority of SaMFormer over representative methods.
\end{abstract}
\section{Introduction}
\label{sec:intro}

People perform dynamic interactions every day, which can be either normal (\eg, handshake, hug and high-five) or abnormal  (\eg, punch, push and rob). Understanding such interactions in video streams is pivotal to downstream applications, such as security surveillance, human-centered video analysis, key event retrieval and sociological investigation \cite{geoSociaRelatPR19}. In this paper, we introduce a novel task termed human-to-human interaction detection (HID) in videos, which aims to 
% obtain a per-frame prediction of all \textbf{I}nteractive \textbf{C}lusters (\textbf{IC}). 
predict all \textbf{I}nteractive \textbf{C}lusters (ICs) per video frame.
% Following the definition of F-formation in sociology \cite{tempEncFormMM13}, 
As defined in F-formation \cite{tempEncFormMM13} in sociology studies,
we assume that different ICs in the same 
% image
scene are non-overlapping, \ie each person belongs to only one IC.
% a person can only belong to one IC. 
Each IC prediction consists of the bounding boxes and the action labels of all its participants. For example, there are two ICs in Figure~\ref{fig:teaser}(b) (one encoded by the green boxes and the other by the red boxes), each of which forms a conversational group.

\noindent\textbf{Why HID?} HID narrows the gap between current  
human interaction understanding methods and the practical demands of downstream applications. 
On the one hand, most existing approaches chose to oversimplify the understanding of human interactions, resulting in either an image/video-level classification task \cite{sd, ip, bit, pam, tvhi, arg, lan2012discriminative, shu2019hierarchical, 2018stagNet, 2020empowerRN, 2017GernShu, 2018Mostafa}, or a problem of just splitting people into multiple groups \cite{statAnaFormbmvc11, tempEncFormMM13, whomDInterICPR16, geoSociaRelatPR19, srrijcv20}. 
Their outputs are too coarse to provide a comprehensive interpretation to human interactions, and cannot answer questions such as \emph{the actions of all participants? the interactive relations among people? and the roles of different subjects?}
On the other, recent methods employ a bottom-up strategy to tackle the problem. First, they detect human bodies with some off-the-shelf detectors \cite{fasterrcnn, fcos, detr}. Second, taking the detected RoIs and the image/video as inputs, a relational model \cite{cagnet, arg, gdn, spatio, wangtip21} is applied to predict the possible interactive configurations (Figure~\ref{fig:teaser}(a)). Such two-stage solutions can only obtain the sub-optimal results on each sub-task, and scales inferior to scenes with many people. Moreover, currently available datasets, including UT \cite{ut}, TVHI \cite{tvhi} and BIT \cite{bit} are designed to understand simple, choreographed human interactions (\eg the \emph{handshaking} depicted by Figure~\ref{fig:teaser}(a)). The lacking of realistic, large-scale and complex datasets profoundly constrains the development and validation of techniques in this field. Because of the aforementioned reasons, we propose {\romannumeral 1}) the HID task, {\romannumeral 2}) an associated challenging benchmark, and {\romannumeral 3}) a strong baseline for HID to facilitate further research in this field.

The HID task is inspired by the well-studied task of \emph{human-object interaction detection} (\textbf{HOID}) \cite{qpic, cdn, ppdm, hotr}, which takes into consideration the target detection, target classification
and the pairwise interactive relation recognition in the form of $\langle$\emph{person ID}, \emph{object ID}, \emph{interactive category}$\rangle$. HID differs from HOID on a particular concentration on human-to-human interactions and the inclusion of high-order interactive relations, which commonly appear in practice. Based on AVA \cite{ava}, which is a benchmark for action detection, we build a large-scale dataset (namely, AVA-I) for HID, in order to mitigate the manifest gap between existing datasets and practical human interactions. We use AVA because it includes abundant daily interactions such as \emph{fight, grab, serve, talk and sing to people}, \etc. More importantly, it contains many challenging cases, \eg, \emph{concurrent interactions}, \emph{heavy occlusions} and \emph{cluttered backgrounds}, which are missing in existing benchmarks. To create AVA-I, we developed a toolbox, which loads and visualizes AVA videos and annotations (including per-person bounding boxes and action labels), and offers the function of adding annotation on interactive relations. In a nutshell, AVA-I includes 85,254 annotated frames, 17 interactive categories and 86,338 interactive groups in total, which significantly surpass existing datasets available to HID (as illustrated by Table~\ref{tab:dataset_comparison}).

To tackle the proposed HID task, we propose a novel one-stage Transformer-based \cite{attention} framework, termed \emph{SaMFormer} (shown in 
 Figure~\ref{fig:SaMFormer}), which predicts human bounding boxes, per-person action labels and interactive relations jointly in a \emph{Split-and-Merging} manner. Specifically, we design two Siamese decoders (sharing a global encoder), \ie, an instance decoder and a group decoder to detect people and interactive groups, respectively (the \emph{split}). Then the model incorporates both spatial cues and semantic representations to associate each human instance to a particular interactive group (the \emph{merging}). We show that SaMFormer is a strong baseline compared with other alternative models. 

To summarize, our contributions are of four aspects: 
\textbf{I)} We propose the HID task, along with a challenging benchmark AVA-I that significantly surpasses existing public ones in complexity and scale.
\textbf{II)} We present a simple encoding-decoding framework SaMFormer for one-stage HID, which admits a holistic understanding of human interactions in a \emph{split-and-merging} way; \textbf{III)} We found that both semantic and spatial cues are crucial to disentangle different interactive groups, and their combination gives the best performance; \textbf{IV)} Without bells and whistles, our proposed SaMFormer achieves leading performance on both existing benchmarks and the proposed AVA-I.

\section{Related Work}
\label{sec:related}

\noindent\textbf{Closely Related Tasks.}
Here we briefly review the most closely related tasks, including action detection (\textbf{AD}) \cite{ava}, human interaction understanding (\textbf{HIU}) \cite{cagnet},  \textbf{HOID} \cite{hoid}, and social relation recognition (\textbf{SRR}) \cite{srrcvpr17, statAnaFormbmvc11}.
AD extends the well-explored object detection task \cite{fasterrcnn, fcos} to the localization of human actions in videos \cite{woo, acrn, ltfbdvu2019, arcnet22}. In general, it solves two sub-problems (either in a joint or split manner), \ie the bounding box regression and the action classification with two predictive heads. The key difference to our HID task is that AD and its associated models are interactive-relation agnostic.
HIU assumes that bounding boxes of human bodies have been obtained beforehand, and performs per-person action recognition and \emph{pairwise} interactive relation estimation simultaneously \cite{cagnet, gdn, spatio, ip, sd, wangtip21}. In comparison, HID takes the prediction of both bounding boxes and \emph{high-order} interactive relations into consideration. We empirically demonstrate that our proposed one-stage framework outperforms recent two-stage HIU models significantly (revealed in Table~\ref{tab:main}).
Compared with HOID, HID focuses specifically on detecting human-to-human interactions such as fighting, chatting, robbing and chasing. Another key difference is that people might have different personal actions (\eg \emph{pushing} \emph{vs.} \emph{falling}) in HID, even if they belong to the identical interactive group (\eg \emph{fighting}).
SRR recognizes the social relations among people appeared in images \cite{srrijcai18, srrijcv20}. Methods in this line rely on the sociological concept ``F-formation'' to disentangle different interactive groups \cite{statAnaFormbmvc11, tempEncFormMM13, whomDInterICPR16, geoSociaRelatPR19}. Compared with HID, SRR aims at assigning a global relational category to an image, which is too coarse to provide a holistic understanding to the scene.

\noindent\textbf{Available Datasets for HID.}
Training HID models requires comprehensive annotations on per-person bounding boxes, per-person actions as well as the interactive relations among people. Available benchmarks, including TVHI \cite{tvhi}, BIT \cite{bit}, UT \cite{ut} and NTU RGB+D \cite{cvpr16nturgbd} \footnote{We would like to note that only a proportion of NTU RGB+D \cite{cvpr16nturgbd} videos include pairwise human-to-human interactions, and this dataset is crafted to support the development of 3D action representation methods using skeletal points.}, are originally crafted for the task of image/video-level interaction classification. Unfortunately, these datasets are of small-scale in terms of the numbers of interactive groups (see Table~\ref{tab:dataset_comparison} for statistics), and only include artificial human interactions performed by a few amateurish actors. Moreover, none of them involve concurrent human interactions (\ie multiple interactions performed by different groups). AVA \cite{ava} is a recently created datasets for spatio-temporal action detection. It is a YouTube-sourced dataset including abundant interactive groups and richer action classes. Also, AVA videos usually include crowded foreground, cluttered background and concurrent interactions. Our AVA-I utilizes all videos and annotations of AVA that contain interested interactions and adds high-quality annotations on interactive relations among people.

\noindent\textbf{Deep Action Representation in Videos.}
The proposed SaMFormer uses deep architectures as the backbone to extract human motion representations. While many effective candidates \cite{3dcnn, c3d, kin, x3d, videomae22, arcnet22} are available, we choose SlowFast \cite{slowfast} for its superior efficacy. SlowFast includes two sideways in architecture, which respectively accept inputs with slower and faster frame rates in order to extract 3D appearance and motion features effectively.

\noindent\textbf{Vision Transformers for Set Prediction.}
The core mechanism of Transformers \cite{attention} is the \emph{self-attention}, which is shown to be advantageous in terms of modeling long-range relations \cite{attention,videotransformer}. In vision, Transformers have been taken to accomplish the so-called \emph{set prediction} tasks, such as object detection \cite{detr,defodetr,dndetr,dabdetr}, instance segmentation \cite{isda, vistr} and HOID \cite{qpic, cdn, hotr, ssrt}, where the final prediction comprises of a set stand-alone components. Though pure Transformers seem promising, most approaches still leverage CNNs to extract base features for attention calculation, in order to avoid the computational overhead of training from scratch. The most closely related work to ours is HOTR \cite{hotr} for HOID, which first predicts sets of instances and interactions, then reconstructs human-object interactions with \emph{human-object Pointers}. Nevertheless, each such pointer is designed to recompose the interactive relation between a person and an object, which is incapable of processing high-order interactions among multiple targets in HID. In comparison, SaMFormer introduces a \emph{split-and-merging} way to associate instances with interactive groups, which enables the prediction of interactions with arbitrary orders.
\section{HID Task}
\label{sec:hid}

\subsection{Problem Definition}
\label{subsec:problem}

Given an input video, HID aims to make each frame a joint prediction on \emph{per-person bounding boxes} $\bm{b} \in  \mathbb{R}^4$, \emph{per-person interactive actions} $\bm{c}\in \{0,1\}^K$ where $K$ is the number of interested interactive actions, and the \emph{interactive relations} among people. Here $\bm{c}$ is a sequence containing $K$ zero-one values, in which $1$ indicates that the corresponding action is activated. Note that  $\bm{c}$ can take multiple $1$s in order to allow the prediction of plural actions performed by the same person (\eg \emph{watching} and \emph{talking}). The interactive relations are denoted by $\mathcal{G}=[g_i]_{i=1}^M$, where $g_i \subseteq \{1,\ldots, N\}$, is a grouping of $N$ people (proposals) into $M$ groups such that $g_i \cap g_j$ = $\emptyset$, $\forall i, j \in \{1,2,\ldots, M\}$, and $i\neq j$. In other words, a proposal can belong to only one group. Moreover, we predict a foreground confidence score $s_i^P$ for each person $i$, and a group confidence score $s_k^G$ for each group $k$. To sum up, each HID prediction takes a form:
\begin{align}
	\bm{y} = [B,C,\mathcal{G},\bm{s}^P,\bm{s}^G],
\end{align}
where $B=[\bm{b}_i]$, $C=[\bm{c}_i]$, $\bm{s}^P=[s^P_i]$, and $\bm{s}^G=[s^G_k]$. Ideally, individuals in a predicted group $g$ have interactive relations with each other (\eg the handshaking and talking scenes in Figure~\ref{fig:teaser}). Note that they might have different personal actions, \eg talking \emph{vs.} listening. Nevertheless, their actions are closely and strongly correlated with each other, thus forming a semantic group.

\subsection{Evaluation Metrics}
\label{metric} 

We use two evaluation metrics for HID. For per-person interactive action detection (a mixture of human bounding box detection and action classification), we use mean average precision ($\rm{mAP}$), which is a popular protocol in AD \cite{ava, woo, slowfast}. For the prediction of interactive relations (\emph{i.e.} $\mathcal{G}$), we design a metric termed \emph{group average precision} (${\rm AP}^G$).  
To this end, we introduce the group-level intersection-over-union ${{\rm IoU}^G}$ between a ground-truth group $g$ and a predicted group $g'$. Let $U=|g|$ and $V=|g'|$.
First, we obtain an \emph{action-agnostic} matching between these two groups with the \emph{optimal bipartite matching} \cite{hungarian}, which relies the cost matrix $O=[ cost_{ij}]_{U \times V} $, with each  $cost_{ij}$ representing the matching cost between two targets $i\in g$ and $j\in g'$:
\begin{align}
	\label{eq:cost_iou}
	&cost_{ij}       =   
	\left\{  
	\begin{array}{ll}
		-{\rm{IoU}}(\boldsymbol{b}^i, \boldsymbol{b}^j) \qquad &{\rm if} \ {\rm{IoU}}(\boldsymbol{b}^i, \boldsymbol{b}^j) \geq  0.5,  \\[1mm]
		\epsilon          \qquad &{\rm otherwise}. \\
	\end{array}
	\right.
\end{align}
Here $\epsilon$ is a big positive value used to reserve matches taking high $\rm{IoU}$ values. Second, taking as input the cost matrix $O$,
a matching candidate between $g$ and $g'$ is computed efficiently utilizing hungarian algorithm \cite{hungarian}. The final matching is obtained by ruling out bad matches having ${\rm{IoU}} < 0.5$.
Finally, the ${{\rm IoU}^G}$ is given by
\begin{align}
	\label{eq:group_iou}
	{{\rm IoU}^G} = \frac{R}{U+V-R},
\end{align}
where $R$ is number of matched elements in $g'$. Like the ${\rm{IoU}}$ in object detection, ${\rm IoU}^G$ measures the rightness of associating the ground-truth $g$ with the prediction $g'$. The ideal case is that $g$ and $g'$ admit a perfect match (\ie $U = V = R$) such that ${\rm IoU}^G=1$. To calculate ${\rm AP}^G$, we use six ${\rm IoU}^G$ thresholds (denoted by $\delta$) ranging from 50\% to 100\% with a resolution of 10\% (similar to COCO detection \cite{coco}). For each $\delta$, the \emph{average precision} (${\rm AP}_{\delta}^G$) is computed under a retrieval scheme of the top-K predicted groups based on ${\rm IoU}^G$ values and $\mathbf{s}^G$ confidences. Finally, ${\rm AP}^G$ is computed by averaging all $\rm{AP}^G_{\delta}$ values.

\subsection{The AVA-Interaction Dataset}
\label{sec:avai}

As mentioned in Section~\ref{sec:related}, available datasets for HID fail to match practical human interactions in terms of scale and complexity. To fix this, we present a large-scale benchmark. Instead of building the benchmark from scratch, the proposed benchmark takes videos from AVA \cite{ava}, which is a large-scale action detection dataset consisting of 437 YouTube videos and 80 atomic daily actions. To make it available to HID, we upgrade the original annotation of AVA by adding frame-by-frame interactive relations among people. Due to this, we call this new version \emph{AVA-Interaction} (AVA-I).

\begin{figure}
	\centering
	\includegraphics[width=0.4\textwidth]{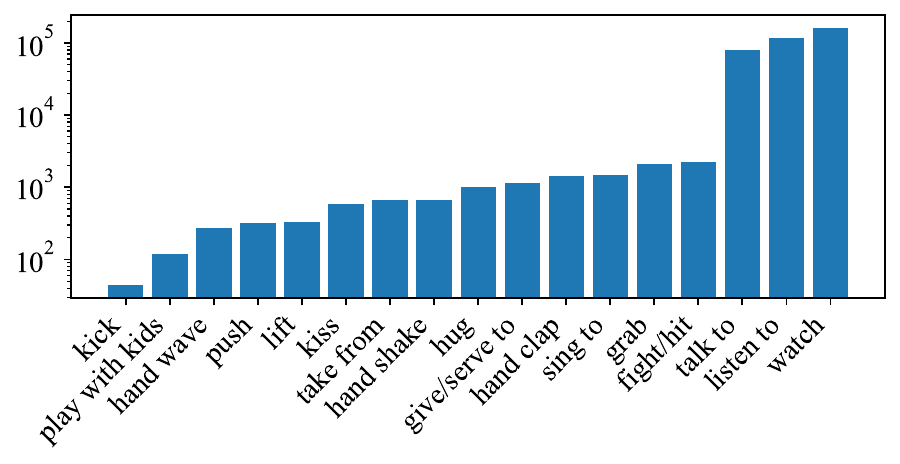}
	\caption{The distribution of interactive classes in AVA-I.}
	\label{fig:label_statistic}
\end{figure}

\begin{figure}
	\centering
	\includegraphics[width=0.4\textwidth]{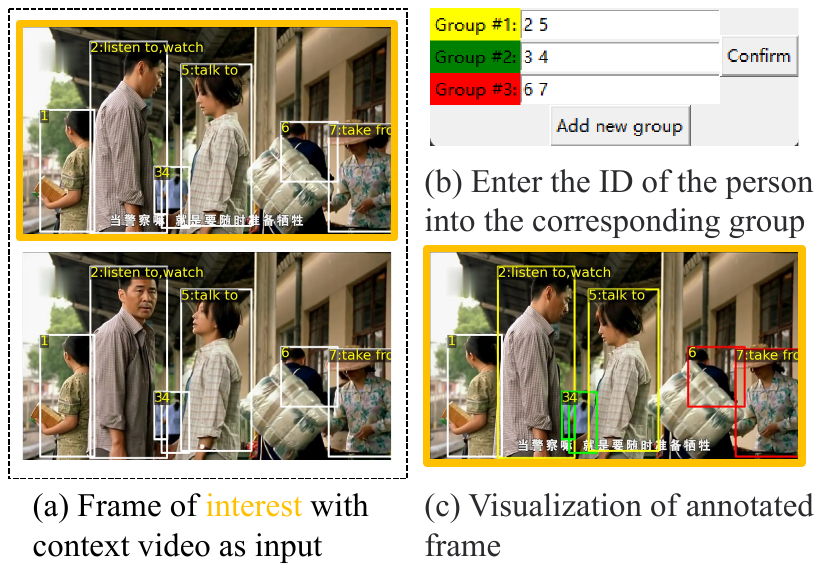}
	\caption{An example of annotating the interactive relations. Per-person action labels and boxes are provided by AVA \cite{ava}.}
	\label{fig:anno}
\end{figure}

\begin{table}
	\centering
        \scriptsize
 	\caption{Compare AVA-I against existing datasets that are available to HID. AVA-I benchmarks HID in terms of all aspects.}
	\begin{tabular}{l|cccc}
		\hline
		& BIT \cite{bit}    & UT \cite{ut}    & TVHI \cite{tvhi}  & AVA-I           \\
		\hline
		\hline
		\# Interaction categories & 9      & 6     & 5     & \textbf{17}  \\
		Max \# group per frame   & 1      & 1     & 1     & \textbf{4}      \\
		Max \# people per group  & 2      & 2     & 2     & \textbf{13}     \\
		Mean \# people per group      & 2      & 2     & 2     & \textbf{2.5}    \\
		\# Annotated frames       & 12,896 & 9,228 & 2,815 & \textbf{85,254} \\
		\# Groups in total        & 12,896 & 9,228 & 2,815 & \textbf{86,338}\\
		\hline
	\end{tabular}%
	\label{tab:dataset_comparison}%
\end{table}%

\begin{figure*}[t!]
	\centering
	\includegraphics[width=.9\textwidth]{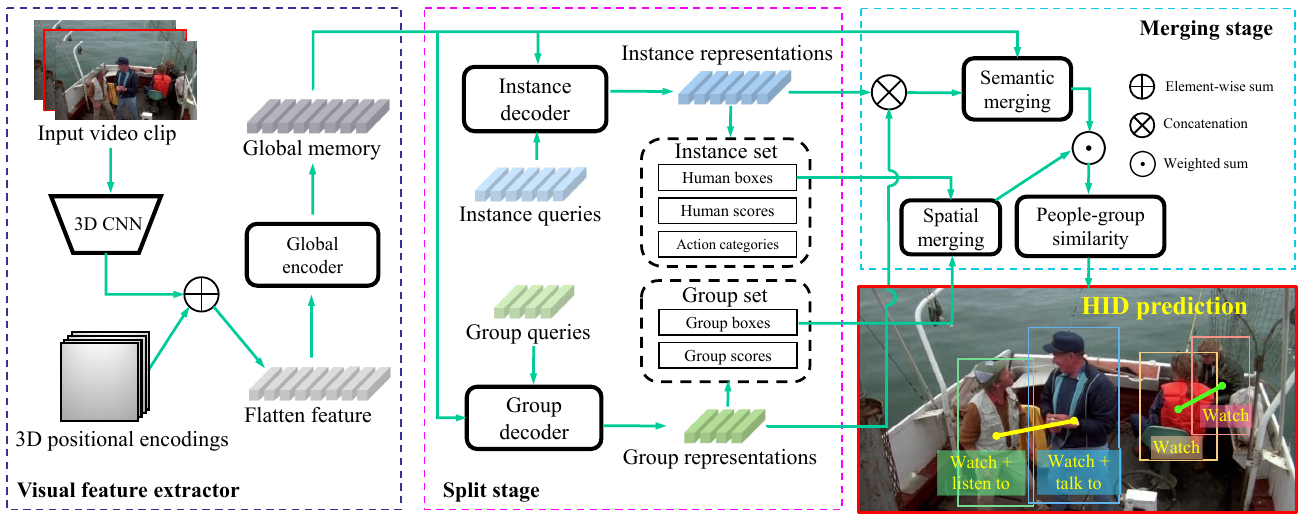}
	\caption{SaMFormer comprises of three components: a \emph{visual feature extractor} (Section~\ref{sec:vfe}) to extract motion features (stored in a global memory) from videos, a \emph{split stage} to predict people and group sets with two decoders (Section~\ref{sec:split}), and a \emph{merging stage} to predict the ownership of people to groups (Section~\ref{sec:merge}). People within the identical group are linked by a line in HID prediction.}
\label{fig:SaMFormer}
\end{figure*}

To shape AVA (created for action detection) for HID, first we discard videos without any human-to-human interactions. Consequently, we obtain 298 videos containing 17 interactive action categories, see Figure~\ref{fig:label_statistic} for statistics. Since AVA does not cover interactive relations among people, we next annotate this information in the frequency of 1 frame per second, which aligns with the original AVA annotation. To this end, we develop an annotation toolbox, which first loads and visualizes an AVA video and its original annotations of bounding boxes and action labels, and then enables the user to browse and annotate the interactive relations in a frame-by-frame manner. It considers three scenarios for efficiency: 1) For frames that only include one person, they are simply discarded as any interaction must includes multiple participants; 2) For frames that exactly include two interactive people (the frame might take non-interactive people as well), the interactive relations are automatically recorded; 3) For all rest scenarios, two groups of human annotators are employed to carefully and independently draft the annotation (Figure~\ref{fig:anno} shows an example of the annotation process). If the two versions are identical, they are believed to be trustworthy and are kept. Otherwise, they are double-checked by a senior researcher in this field to finalize the annotation. In this way, the quality of the annotated interactive relations can be guaranteed.

The comparison of AVA-I against existing datasets is shown in Table~\ref{tab:dataset_comparison}. AVA-I outperforms existing datasets in terms of the numbers of annotated frames, the total interactive groups, and the interaction classes. AVA-I is also the best regarding the largest numbers of people and interactive groups per frame, as well as the average number of people per group. Unlike BIT and UT (acted and captured by amateurish actors and videographers), AVA-I examples are sourced from TV-episodes, which renders it highly proximate to daily life.
\section{SaMFormer}
\label{sec:hidetr}
The architecture of our proposed SaMFormer is shown in Figure~\ref{fig:SaMFormer}. Given a video clip, we first apply a \emph{Visual Feature Extractor} (VFE, in Section~\ref{sec:vfe}) to extract spatio-temporal visual features, which are stored in the global memory for subsequent usage. Next, taking as inputs the features in the global memory, the instance queries, and the group queries, we utilize two parallel Transformer decoders (\ie, the instance decoder and the group decoder) to query instances (people) and interactive groups (\ie the \emph{Split} stage, in Section~\ref{sec:split}). Then the estimated instances and groups are associated with each other to generate the final HID prediction (\ie the \emph{merging} stage, in Section~\ref{sec:merge}). Below we elaborate each component.

\subsection{Visual Feature Extractor}
\label{sec:vfe}
The VFE combines a 3D CNN backbone \cite{slowfast} and a Transformer encoder. 
Feeding with an input video clip $V\in \mathbb{R}^{T\times H_0\times W_0\times 3}$, the backbone generates a feature map as a tensor, taking the shape $(T, H, W, D)$, where $T, H, W, D$ denote the number of frames, the height, the width and the depth of the feature map respectively.
%
% below can be used in the journal version
%
%\red{First, assume the input video tensor $V$ has a shape of $(T, H_0, W_0, 3)$, where $T$, $H_0$ and $W_0$ represent the number of frames, and the original height and width of the input frame, respectively. The backbone takes $V$ as input and generates feature map with a shape of $(T, H, W, D)$, where $H$, $W$, and $D$ represent the downsampled height, width, and number of channels, respectively.}
%
Then the depth of the feature map is reduced to $d$ (\eg 256) by applying an $1\times1\times1$ filter. Next, a flatten operation is taken to get the feature $\bm{x} \in \mathbb{R}^{(T \cdot H \cdot W) \times d }$ by collapsing the temporal and spatial dimensions.
Since the Transformer architecture is permutation-invariant, we supplement $\bm{x}$ with fixed 3D positional encodings $\bm{e} \in \mathbb{R}^{(T \cdot H \cdot W) \times d}$ \cite{attention, vistr}, and feed them into a Transformer encoder. Benefiting from the attention mechanism \cite{attention}, the encoder injects rich contextual information into $\bm{x}$, and the resulting representations $\bm{x}_{g} \in \mathbb{R}^{(T \cdot H \cdot W) \times d}$ are stored into a global memory for subsequent usage.

\subsection{The Split Stage}
\label{sec:split}
This stage is designed to predict the instance and group sets in parallel. To this end, we first randomly initialize two sets of learnable queries, \ie, $\bm{Q}^{P}\in \mathbb{R}^{u  \times d}$ as instance queries, and $\bm{Q}^{G } \in  \mathbb{R}^{v \times d}$ as group queries, where $u$ and $v$ denote the allowed largest numbers of instances and groups. Then these queries, together with $\bm{x}_{g}$ (served as key and value in cross-attention), are fed into the  instance decoder and the group decoder as illustrated by Figure~\ref{fig:SaMFormer}. The instance decoder transforms the instance queries into instance representations $\{\bm{r}^{P}_i\}_{i=1}^{u}$ for the detection of person proposals and the recognition of per-person actions, while the group decoder leverages the group queries to extract group representations $\{\bm{r}^{G}_j\}_{j=1}^{v}$ for the detection of interactive groups. We then apply three shared MLPs to each instance representation $\bm{r}^P_i$, which respectively predict a confidence score $s^P_i$, an interactive action category $c_i$ and a bounding box $\bm{b}^P_i$ for the $i$-th instance query. Likewise, we apply two extra FFNs to each group representation $\bm{r}^G_j$, in order to predict a group confidence score $s^G_j$ and a group bounding box $\bm{b}^G_j$ (\ie a bounding box enclosing all interactive people in this group). Consequently, we obtain an estimated instance set $ \{s^P_i, c_i, \bm{b}^P_i \}_{i=1}^{u}$ using the instance decoder, and an estimated group set $\{s^G_j, \bm{b}^G_j\}_{j=1}^{v}$ with the group decoder.

\subsection{The Merging Stage}
\label{sec:merge}
The goal of this merging stage is to merge the output from the previous split stage to predict the interactive relationships. Specifically, we introduce two forms of merging: semantic merging and spatial merging. For the former, we take the instance representations $\{\bm{r}^{P}_i\}_{i=1}^{u}$ and group representations $\{\bm{r}^{G}_j\}_{j=1}^{v}$ output by the decoders to compute the people-group similarity $\Theta = [\theta_{ij}]_{u \times v}$ (\ie the \emph{semantic similarity}), with ${\theta}_{ij}=1$ indicating that person $i$ definitely belongs to group $j$. We implement this with three alternatives.

\noindent\textbf{Inner Product} performs the inner (dot) product between linearly transformed instance and group representations:
\begin{align}
	\theta_{ij} = \frac{\langle f^P(\bm{r}^P_i), f^G(\bm{r}^{G}_j)\rangle}{\left\lVert f^P(\bm{r}^P_i) \right\rVert \cdot \left\lVert f^G(\bm{r}^{G}_j) \right\rVert},
\end{align}
where $f^P(\cdot)$ and $f^G(\cdot)$ are linear transformations, $\langle\cdot,\cdot\rangle$ denotes inner product, and $\left\lVert \cdot \right\rVert$ is $L2$ normalization.

\noindent\textbf{Linear Transformation} follows \cite{gat}, which first concatenates instance and group representations. Then it applies a linear transformation to the concatenated feature:
\begin{align}
	\theta_{ij} = {\rm sigmoid}\big(f^C(\bm{r}^P_i \otimes \bm{r}^{G}_j)\big),
\end{align}
where $\otimes$ denotes vector concatenation.

\begin{figure}[t!]
	\centering
	\includegraphics[width=.92\linewidth]{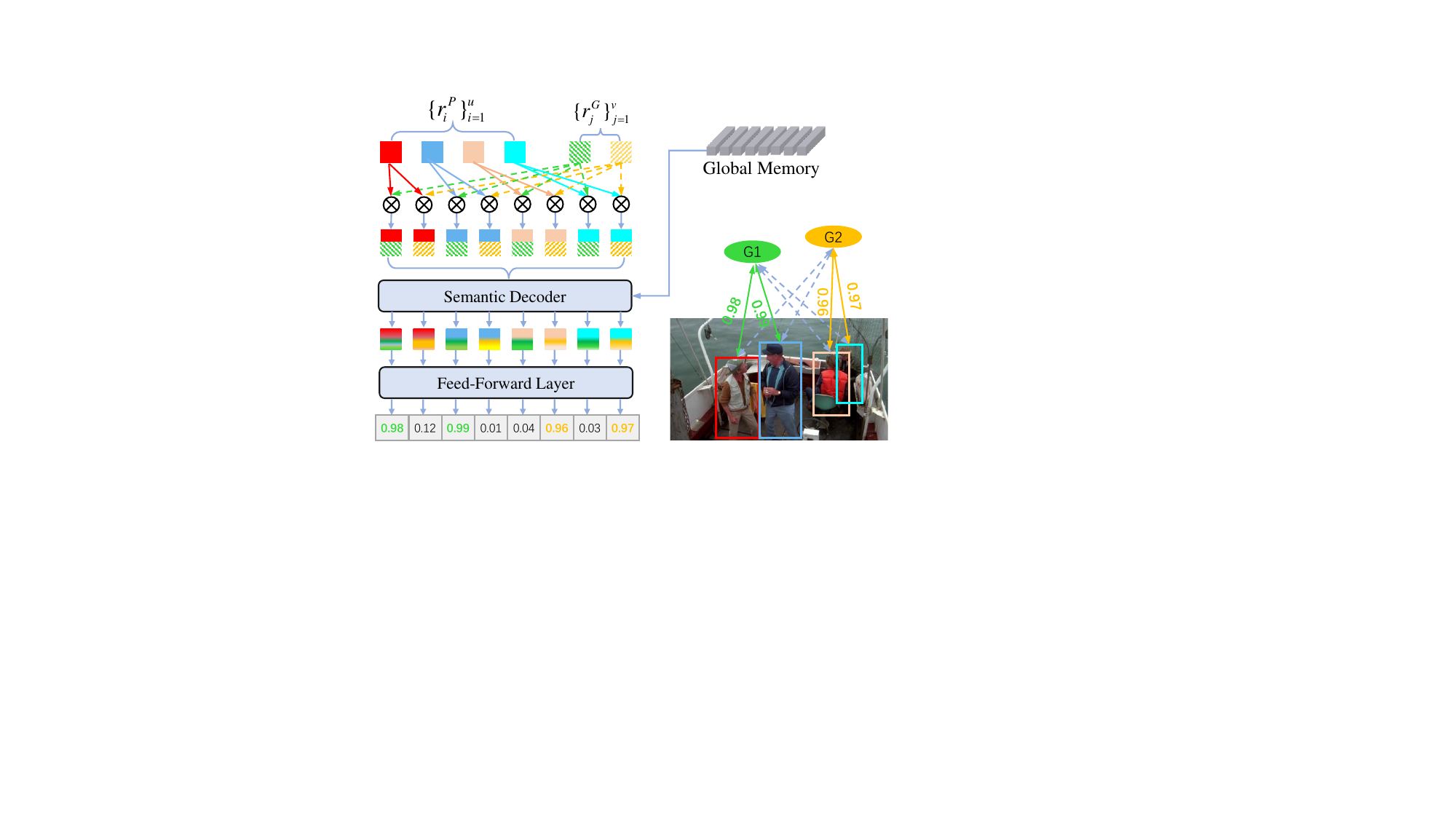}
	\caption{The semantic decoder. Here G1 and G2 denote two group candidates. Best viewed in color.}
	\label{fig:semantic_decoder}
\end{figure}

\begin{figure}
	\centering
        \includegraphics[width=.37\textwidth]{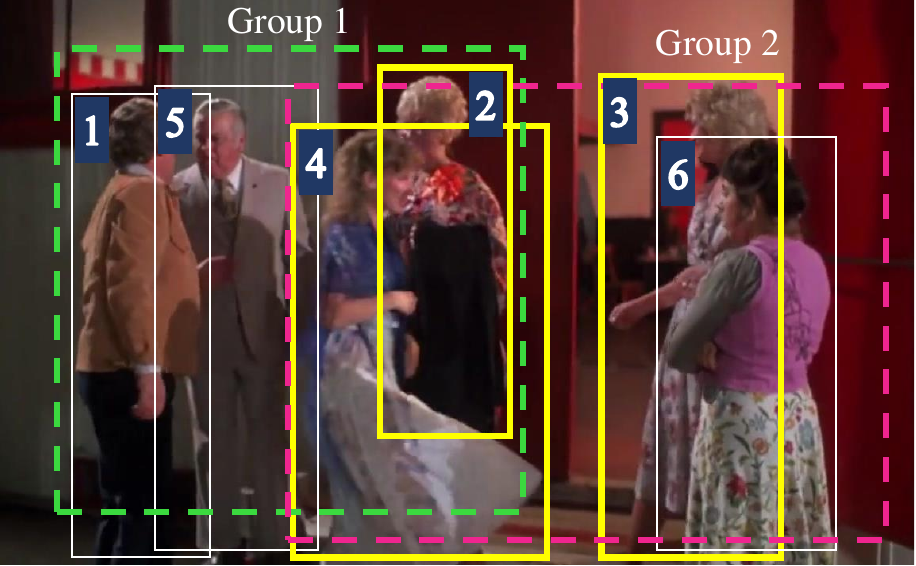} \label{fig:top_image_a}
                \hspace{-.5cm}
        \includegraphics[width=.4\textwidth]{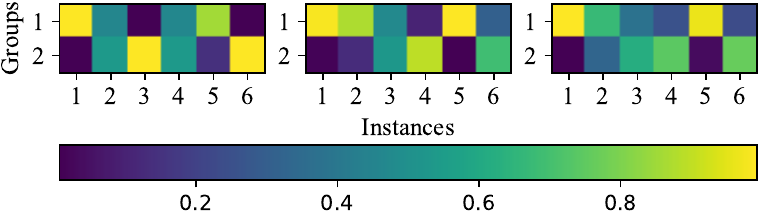} \label{fig:top_image_b}
	\caption{Grouping instances according to their interactive relations requires both spatial and semantic cues. Top: an image with detected instances (marked by boxes with solid edges and IDs) and groups (dashed boxes). Yellow boxes highlight hard instances in terms of grouping; Bottom: The predicted $\Theta$ matrices corresponding to the spatial similarity (left), the semantic similarity (middle) and their blend (right). Pure spatial or semantic cue could be insufficient in terms of grouping tricky instances, \eg  person~\#2 (against group 2 due to her proximity to person~\#4), \#3 (against group 1 due to her head orientation) and \#4 (against group 1 due to her proximity to person~\#2). Overall the blending version is more robust in tackling such challenges.}
	\label{fig:iou_and_semantic}
\end{figure}

\noindent\textbf{Semantic Decoder.} Both the \emph{inner product} and the \emph{linear transformation} have omitted the global memory, which might contain key contextual information beyond the scope of local features in terms of discriminating interactive relations. Motivated by this, we design an additional Transformer decoder named \emph{semantic decoder} (illustrated by Figure~\ref{fig:semantic_decoder}), which takes $[f^R(\bm{r}^P_i\|\bm{r}^G_j)]_{i=1,\ldots, u, j = 1,\ldots, v}$ as queries, and $\bm{x}_g$ as keys and values. Here $f^R(\cdot)$ maps each concatenated vector to a new vector in $\mathbb{R}^d$. Let $\bm{r}^{PG}_{ij}\in \mathbb{R}^{d}$ be the output of the decoder for the person $i$ and the group $j$. Then a FFN (shared by all $\bm{r}^{PG}_{ij}$ vectors) is taken to compute the semantic similarity $\theta_{ij}$.

Although the \emph{semantic merging} block is capable of discriminating most interactive relations, it could fail under some tricky circumstances, see Figure~\ref{fig:iou_and_semantic} for an interesting example. To alleviate this, we introduce a spatial prior, which is encoded by another $\Theta' = [\theta'_{ij}]_{u \times v}$ matrix with entries given by:
\begin{align}
	\theta'_{ij} = {\rm IoF} (\bm{b}^P_i, \bm{b}^G_j),
\end{align}
where ${\rm IoF}(\cdot, \cdot)$ computes \emph{intersection-over-foreground}, and the person-box is used as the foreground. Specifically, if the person-box $i$ is thoroughly enclosed by the group-box $j$, then ${\theta}'_{ij}=1$. We name grouping with $\theta'$ the \emph{spatial merging}. We get the final similarity by blending spatial and semantic similarities:
\begin{align}
\label{eq:alpha}
	\hat{\theta}_{ij} = \alpha  \theta'_{ij} + (1 - \alpha)\theta_{ij}, \quad \alpha \in [0,1].
\end{align}
Here, $\alpha$ can be pre-designated empirically or learned from data. Figure~\ref{fig:iou_and_semantic} demonstrates that such blending with $\hat{\theta}$ is more effective in terms of disentangling complicated human interactions.

\subsection{Training and Inference}
\noindent\textbf{Training.}
Following the training protocol of DETR \cite{detr}, we first match each instance or group ground truth with its best-matching prediction with the Hungarian algorithm \cite{hungarian}. These matches are then utilized to calculate losses for back-propagation. The total loss comprises of four terms:
\begin{align}
	L = L^P_{detr} + L^G_{detr}  + \lambda_a^P L_a^P +\lambda^{PG} L^{PG},
\end{align}
%\begin{align}
%	L = (\lambda_b^P L^P_b + \lambda^P_{GIoU} L^P_{GIoU} + \lambda^P_f L^P_f+\lambda_a^P L_a^P) +\notag\\
%	(\lambda_b^G L^G_b+\lambda^G_{GIoU} L^G_{GIoU}+\lambda^G_f L^G_f)+\lambda^{PG} L^{PG},
%\end{align}
where $L^P_{detr}$ and $L^G_{detr}$ are of DETR losses (we follow the loss items and loss weights in DETR, please refer to \cite{detr}), used to supervise foreground-background classification and bounding box regression of people and groups respectively. $L^P_a$ and $L^{PG}$ are cross-entropy losses to supervise per-person action classification and instance-group matching, and $\lambda$ weights different terms.

\noindent\textbf{Inference}
is to recognize all instances and associate them to form proper groups. We first get all instances from the instance decoder. Next, given the similarity matrix $\hat{\Theta} = [\hat{\theta}_{ij}]_{u \times v}$ (with entries computed by Equation~\ref{eq:alpha}), we compute the group ID via $\mathop{\mathrm{argmax}}\nolimits_{k\in\{1,\ldots, v\}}{\hat{\theta}}_{ik}$, for any instance $i$. This operation guarantees that each instance is assigned a unique group ID to meet our grouping assumption.

\section{Experiments}
\label{sec:experiments}
% In this section, we conduct experiments to verify the effectiveness of proposed HiDETR. In Section~\ref{sec:implementation}, we introduce the implementation details of HiDETR, and then several baselines in HID will be claimed in Section~\ref{sec:baselines}. In addition, we

\begin{table*}[htbp]
	\centering
        \small
	\begin{tabular}{l| l lcccccc}
		\toprule
                Type       & Method   & Backbone     & E2E    & $T \times \tau$ & AP$^G$   & AP$^G_{50}$ & AP$^G_{80}$ & AP$^P_{50}$   \\
            \midrule
    \multirow{2}[0]{*}{Action Only} & SlowOnly-R50 & SlowOnly-R50 & \XSolidBrush & $4 \times 16$   & -              & -                 & -                 & 17.05  \\
                & SlowFast-R50 & SlowFast-R50 & \XSolidBrush & $8 \times 8$    & -              & -                 & -                 & 18.06  \\
            \midrule
    \multirow{4}[0]{*}{Two-Stage} 
                & SCG \cite{scg}         & SlowOnly-R50 & \XSolidBrush           & $4 \times 16$       & 59.45    &  73.68       &  51.34       & 17.12  \\
                & SCG \cite{scg} (\textbf{ground truth boxes})        & SlowOnly-R50 & \XSolidBrush      &  $4 \times 16$    & 73.88    & 84.19    & 67.19    &  \underline{19.52}  \\
                & CAGNet \cite{cagnet}       & SlowOnly-R50  & \XSolidBrush             & $4 \times 16$    & 57.11          & 70.77             & 50.85             & 15.08  \\
                & CAGNet \cite{cagnet} (\textbf{ground truth boxes})     & SlowOnly-R50   & \XSolidBrush             & $4 \times 16$            & \textbf{76.62}       & 86.2              & \textbf{72.2}      & 18.36  \\
            \midrule
    \multirow{7}[0]{*}{One-Stage} 
                & QPIC \cite{qpic}          & SlowOnly-R50 & \XSolidBrush           &  $4 \times 16$           &  63.74      &  84.23     & 53.91   & 16.71   \\
                & CDN  \cite{cdn}          & SlowOnly-R50 & \XSolidBrush           &  $4 \times 16$           &   63.81      &  85.26     & 53.27   & 17.59 \\
                \cmidrule{2-9}                & \multicolumn{8}{c}{SaMFormer (ours)} \\

            & Spatial only                        & SlowOnly-R50 & \Checkmark   & $4 \times 16$   & 56.1           & 79.38             & 47.29             & 17.88            \bigstrut[t] \\
            & Semantic only                        & SlowOnly-R50 & \Checkmark   & $4 \times 16$   & 67.14          & 87.53             & 60.13             & 18.01                         \\
            & Spatial + Semantic                          & SlowOnly-R50 & \Checkmark   & $4 \times 16$   & 72.52          & \underline{89.11}             & 66.09             & 18.46                        \\
            & Spatial + Semantic                          & SlowFast-R50 & \Checkmark   & $8 \times 8$    & \underline{74.05} & \textbf{89.12}             & \underline{68.58}             & \textbf{19.97}      \\
		\bottomrule
	\end{tabular}%
	\caption{Comparing SaMFormers with other baselines on AVA-I. E2E means end-to-end trainable. $T \times \tau$ tells the number of frames ($T$) and the sampling-rate ($\tau$) on each input video. For all two-stage methods, Faster R-CNN \cite{fasterrcnn} is taken to detect human bodies as RoIs. \emph{Ground truth boxes} indicate that the annotated bounding boxes are used instead of the detected ones.}
	\label{tab:main}%
\end{table*}%

\noindent\textbf{Implementation Details.}
We use AdamW\cite{adamw} with a weight decay of 0.0001 as the optimizer for all experiments. The mini-batch consists of 24 video clips and all models are trained with 3 pieces of RTX 3090 cards (8 clips on each card). We train the network in 20 epochs with an initial learning rate of 0.0001. The decay factor is 0.1 applied at epoch 10 and 16, respectively. The backbone is initialized with the pre-trained weights on Kinetics \cite{kinetics} and the rest layers are initialized with Xavier \cite{xavier}. We resize each frame to $256 \times 256$ for all methods in both training and evaluation. Following \cite{detr}, both the encoder and decoder of our proposed SaMFormer include 6 layers, and the numbers of both instance and group queries are 30. 
For the loss weights, we use $\lambda_a^P = 1$ and $\lambda^{PG} = 5$, without meticulously tuning.
% \red{Regarding the loss weights, we followed DETR \cite{detr} in $L^P_{detr}$ and $L^G_{detr}$. For the $\lambda_a^P$ and $\lambda^{PG}$, we set them to 1 and 5 respectively, without meticulously tuning.}
For efficient exploration, we use the lightweight SlowOnly \cite{slowfast} as the backbone for the ablation study. We also adopt SlowFast as the backbone for fair comparison against SOTA results.

\noindent\textbf{Dataset and Evaluation Metrics.}
Following AVA \cite{ava}, we split AVA-I into training and testing sets, which contain 234 (66,285 frames) and 64 (18,969 frames) videos respectively. 
We use ${\rm AP}^G$, ${\rm AP}^G_{50}$, ${\rm AP}^G_{80}$ to evaluate the performance on interactive relation detection. For human detection and action classification, we take the mean average precision denoted by ${\rm AP}^P_{50}$ (where the IoU threshold is 50\% as suggested in \cite{ava}).

\vspace{-.2cm}
\subsection{Main Results on AVA-I}
\label{sec:main_results}

As HID is new, we absorb ideas from relevant tasks \cite{cagnet,ava,slowfast,hoid} to create several baselines, which tackle HID in either two-stage or one-stage. The former detects human bodies in the first stage (employing Faster-RCNN \cite{fasterrcnn}), and then performs interaction recognition in the second stage, while the latter accomplishes both sub-tasks in a unique model.
% Two-stage
For the two-stage methods, we choose the CAGNet \cite{cagnet} for HIU and SCG \cite{scg} for HOID as our competitors. As they take external bounding boxes as inputs, we evaluate them using both detected and annotated boxes. Note that both of them can only predict pairwise interactive relations. In order to translate them into $\mathcal{G}$ predictions, we resort to a greedy search: if the interactive score of a pair of people $\geq 0.6$,  they are grouped together. 
% One-stage
For the one-stage approaches, we choose the classic QPIC \cite{qpic} and CDN \cite{cdn} for HOID. 
We utilize them to predict a set of human interaction pairs, which consist of the bounding boxes of two interactive targets and their corresponding action categories. Before doing the aforementioned greedy search, an extra post-processing is performed in order to match the same human instance across different predicted interaction pairs. To this end, we set a threshold (\ie $0.8$), and consider that the two bounding boxes belong to the identical person if the IoU of them is greater than this value. 
We would like to note that our SaMFormer is also of one-stage, but what sets it apart from other one-stage methods is that SaMFormer can achieve end-to-end discovery of interactive groups, without the aid of heuristic search.

Furthermore, we also apply SlowFast/SlowOnly \cite{slowfast} to detect person-wise actions on AVA-I, which serves as a strong baseline for both human detection and per-person action recognition. To understand the contribution of different building blocks of SaMFormer, we test its four variants depending on which backbones are incorporated (\ie SlowFast and SlowOnly), and the particular fusion strategy (\ie \emph{spatial only}, \emph{semantic only} and \emph{spatial + semantic}) to take. Both backbones are based on ResNet50 (R50) \cite{resnet}.

Table~\ref{tab:main} gives the main quantitative results. Thanks to the self-attention modelling and the effective multi-task learning, SaMFormer surpasses the baseline of AD (Action Only), by 1.41 points under SlowOnly and 1.91 points under SlowFast on AP$^P_{50}$.
For interaction detection, SaMFormer (Spatial + Semantic) outperforms both one-stage and two-stage baselines by large margins (except for models using \emph{ground-truth bounding boxes}) under all metrics. When utilizing an out-of-box detector \cite{fasterrcnn} to provide RoIs, the performance of the two-stage models is severely degraded (-14.43 and -19.51 points in term of AP$^G$ for SCG and CAGNet), which indicates that these models typically suffer from the quality of detection. One-stage models jointly optimize three sub-tasks, including human detection, per-person action recognition and interactive relation estimation, and their results are generally better than two-stage methods, which is consistent with the observations in HOID\cite{cdn, qpic, hoid}. Apart from the one-stage design, SaMFormer also leverages a simple yet effective Split-and-Merging paradigm to achieve detection and grouping, which in turn removes the heuristic post-processing procedure (\ie the \emph{greedy search} to form groups). With this, SaMFormer surpasses the second best CDN \cite{cdn} by 8.71 points (72.52 \emph{vs.} 63.81).

Furthermore, we investigate the effectiveness of our merging design. The necessity of combing the \emph{semantic} and the \emph{spatial} merging could be justified by that the conjunctive usage of them improves the result by 5.38 points (last four rows in Table~\ref{tab:main}), which suggests that both the semantic feature and the spatial prior are essential to the estimation of interactive relations. In addition, semantic merging is more reliable than spatial merging, as spatial cues are typically non-robust against dynamic human interactions. Replacing SlowOnly with SlowFast, a more powerful backbone for video-level representation, the result is further improved by 1.53, which demonstrates that SaMFormer largely benefits from our \emph{split} and \emph{merging} designs.

\vspace{-0.2cm}
\subsection{Ablative Study}
\label{sec:ablation}
We take an ablation to validate our model design. Here all methods use SlowOnly-R50 as the backbone for fairness and effectiveness.

\begin{figure}
	\centering
	\includegraphics[width=.37\textwidth]{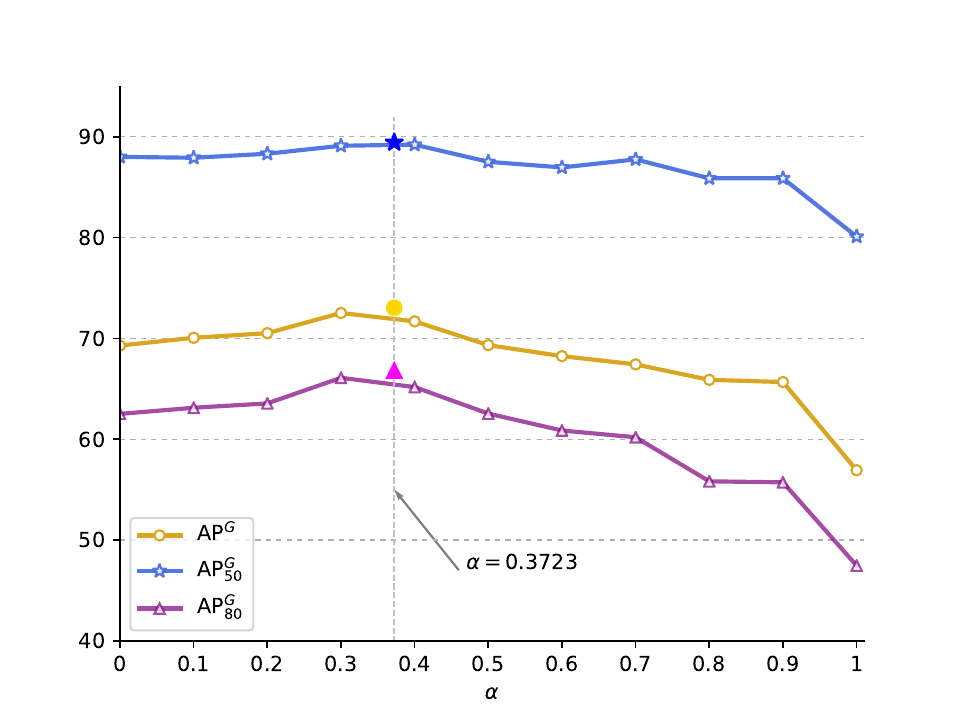}
	\caption{Performance comparison with both pre-designated (lines) and learned (scattered markers) $\alpha$s. Results of AP$^G$, AP$^G_{50}$ and AP$^G_{80}$ correspond to orange, blue and magenta lines (or markers), respectively. Learned $\alpha$ performs marginally better.}
	\label{fig:alpha}
\end{figure}

\noindent\textbf{Spatial \emph{vs.} Semantic.}
SaMFormer incorporates both spatial prior and learned semantic representations to discriminate the interactive relations in the \emph{merging} stage. To ablate such design, we evaluate SaMFormer with both pre-designated and learned $\alpha$s. Figure~\ref{fig:alpha} illustrates the changes on performance when increasing the value of $\alpha$ from 0 to 1. One can see that a proper blending (\ie $0.1\leq \alpha\leq 0.4$) is more favourable than merely using spatial or semantic information (\ie $\alpha=0$ or $\alpha=1$). When using pre-designated values, the best result on ${\rm AP}^G$ is obtained when $\alpha = 0.3$, while a further increasing of it degrades the performance. Hence we conclude that the learned representation is more reliable than the spatial heuristic. Overall, the learned $\alpha$ (0.3723) is only marginally better than the pre-designated value (0.3), which suggests that the blending is insensitive to the perturbation on $\alpha$. Table~\ref{tab:spatial_semantic} gives the detailed results on this comparison.

\begin{table}[t!]
	\centering
        \small
		\begin{tabular}{ll|cccc}
			\toprule
			Type & $\alpha$ & AP$^G$   & AP$^G_{50}$ & AP$^G_{80}$ & AP$^P_{50}$        \\
			\midrule
                Pre-designated        & 0.3         & 72.52       & 89.11       & 66.09       & \textbf{18.46} \\
                Learned   & 0.3723      & \textbf{73.04}       & \textbf{89.44}       & \textbf{66.81}  & 18.42 \\
                \bottomrule
		\end{tabular}%
	\caption{Compare the performance of learned $\alpha$ with that of pre-designated $\alpha$ (\ie $\alpha=0.3$). Learned $\alpha$ is marginally better.}
	\label{tab:spatial_semantic}%
\end{table}%

\begin{table}[tbp]
	\centering
        \small
	\begin{tabular}{l|cccc}
		\toprule
		Methods               & AP$^G$   & AP$^G_{50}$ & AP$^G_{80}$ & AP$^P_{50}$ \\
		\midrule
		Linear Transformation & 70.83          & 87.44             & 64.97             & 17.82             \\
		Inner Product         & 70.92          & 87.15             & 65.15             & 17.79             \\
  	Semantic Decoder      & \textbf{72.52} & \textbf{89.11}    & \textbf{66.09}    & \textbf{18.46} \\
		\bottomrule
	\end{tabular}
	\caption{Performance comparison on \emph{semantic merging} variants.}
	\label{tab:semantic}%
\end{table}%

\noindent\textbf{Variants of Semantic Merging.}
Section~\ref{sec:merge} presents three alternatives for semantic merging, which are compared in Table~\ref{tab:semantic}. \emph{Semantic Decoder} is notably better than \emph{Linear Transformation} and \emph{Inner Product}. Indeed, the decoder allows to attentively learn informative local and contextual representations for each \emph{person-group} query, such that the interactive relations among people could be determined under a group perspective. In comparison, the other merging methods compute \emph{person-group} similarities under a local perspective, which is insufficient in addressing complicated human interactions.

\subsection{Qualitative Results}
\label{sec:qualitative}
\begin{figure*}[t!]
	\centering
	\includegraphics[width=.91\textwidth]{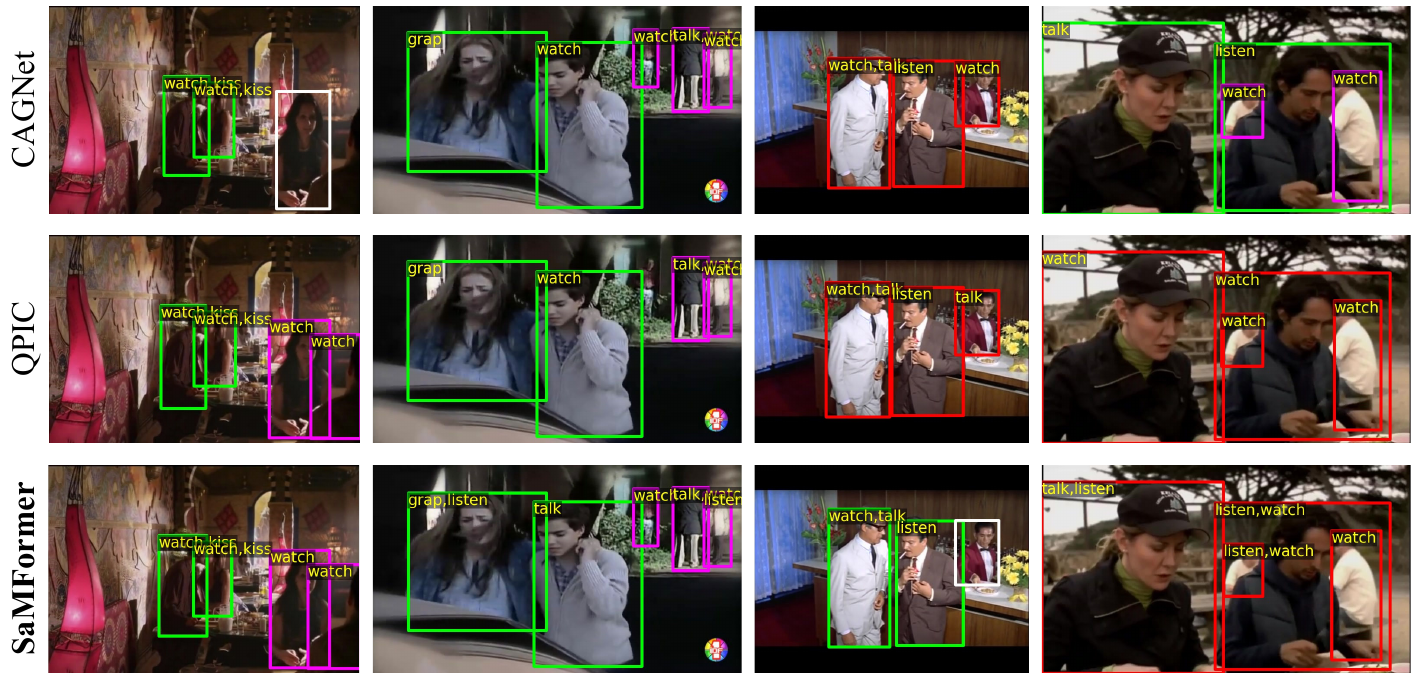}
        \caption{Visualize HID results predicted by different models on AVA-I (best viewed in color, and by zooming in). Each column shows the same example predicted by CAGNet\cite{cagnet} (two-stage), QPIC \cite{qpic} (one-stage) and SaMFormer, respectively. Separate groups are encoded by different colors, with red highlighting incorrect grouping results.}
	\label{fig:qualitative}
\end{figure*}

To provide a qualitative analysis of different models, we visualize a few predictions in Figure~\ref{fig:qualitative}. Here we select CAGNet (using Faster R-CNN to detect human boxes) and QPIC (of one-stage) as the competitors. The three leftmost examples show that CAGNet and QPIC could give defective HID results, either caused by missing detection (the two leftmost columns) or incorrect grouping (the third column from left to right), while our proposed SaMFormer is able to generate correct HID predictions. We attribute this success to the split-and-merging design of SaMFormer, which enables the model to interpret the scene in both micro (\ie instances) and macro (\ie groups) perspectives. However, for the rightmost example, where severe occlusion appears, CAGNet performs much better than QPIC and our SaMFormer. Specifically, QPIC and SaMFormer mistakenly put all targets into an identical group, probably caused by the partial shared representations and the proximity between occluded people.

\subsection{Evaluation on BIT and UT}
\label{sec:ava_and_hiu}

\begin{table}[htbp]
	\centering
	\small
	\begin{tabular}{l|cll}
		\toprule
		Model & Detector & BIT \cite{bit} & UT \cite{ut} \\
		\midrule
		CAGNet \cite{cagnet}                           & YOLO v5 \cite{yolov5}         & 59.39 (+17.22)           & 75.63 (+10.5)          \\
		CAGNet \cite{cagnet}                           & FCOS \cite{fcos}              & 68.62 (+7.99)            & 81.85 (+4.28)          \\
		CAGNet \cite{cagnet}                           & Annotated                 & \textbf{81.73} (-5.12) & 86.03 (+0.1)   \\
		SaMFormer & -                            & 76.61                              & \textbf{86.13}      \\
		\bottomrule            
	\end{tabular}
	\caption{Comparison with CAGNet \cite{cagnet} on BIT \cite{bit} and UT \cite{ut}. SaMformer achieves outstanding performance, which even surpasses CAGNet using annotated bounding boxes on UT. As a HID task is considered, the results of CAGNet here are different from the HIU results presented in \cite{cagnet}.}
	\label{tab:hiu}%
\end{table}%

Table~\ref{tab:hiu} compares SaMFormer with CAGNet \cite{cagnet} on BIT and UT. Since CAGNet takes as inputs the bounding boxes, for fair comparison, we train three CAGNet models using YOLOv5 \cite{yolov5} detections, FCOS \cite{fcos} detections and annotated boxes (testing uses the identical source of boxes). On BIT, SaMFormer outperforms by 17.22 and 7.99 points in terms of mAP against CAGNet with YOLOv5 and FCOS detections respectively. On UT, SaMFormer surpasses CAGNet in all settings, and to our surprise, it is even marginally better than CAGNet with \emph{annotated} boxes.

\section{Conclusion}
\label{sec:conclusion}
We presented the HID task, which devotes to locate people and predict their actions and interactive relations in videos.
Compared with other Siamese tasks like AD and HIU, HID is more comprehensive in terms of analyzing human activities.
Along with HID, a large-scale, realistic and challenging dataset was proposed to benchmark HID, on which we observed salient performance gaps against available datasets with strong baselines. To alleviate this, we proposed SaMFormer based on the encoding-decoding mechanism of Transformer, which allows tackling HID in an end-to-end \emph{Split-and-Merging} manner. Without bells and whistles, SaMFormer outperforms strong baselines by clear margins on both existing and new benchmarks. Proper ablation experiments are also conducted, showing that a proper blending of semantic and spatial cues is vital to HID. Nevertheless, failure cases indicate that SaMFormer struggles to deal with heavy occlusions and suspicious interactions. We hope the proposed new task, dataset and baseline can stir more research in this field.

{\small
\bibliographystyle{ieee_fullname}
\bibliography{_main}
}

% \ifarxiv \clearpage \input{./section/12_appendix} \fi

\end{document}